\documentclass{article}
\usepackage{ijcai19}

\usepackage{times}
\usepackage{soul}
\usepackage{url}
\usepackage[hidelinks]{hyperref}
\usepackage[utf8]{inputenc}
\usepackage[small]{caption}
\usepackage{graphicx}
\usepackage{amsmath}
\usepackage{booktabs}
\usepackage{algorithm}
\usepackage{algorithmic}
\usepackage{amssymb}
\usepackage{color}
\usepackage{tikz}
\definecolor{light-blue}{rgb}{.6, .6, 1}
\definecolor{light-brown}{rgb}{.8, .4, .3}
\definecolor{light-yellow}{rgb}{1, .8, .5}
\definecolor{light-green}{rgb}{.7, 1, .7}

\usepackage{tabularx}
\newcolumntype{M}[1]{>{\centering\arraybackslash}m{#1}}
\newcolumntype{N}{@{}m{0pt}@{}}
\usepackage{float}

\newcommand\A{\mathcal{A}}
\newcommand\SSS{\mathcal{S}}
\newcommand\OO{\mathcal{O}}
\newcommand\M{\mathcal{M}}
\newcommand\R{\mathcal{R}}
\newcommand\IfO{\mbox{\emph{IfO}}}
\newcommand\IL{\mbox{\emph{IL}}}
\newcommand\RL{\mbox{\emph{RL}}}
\newcommand\GAIL{\mbox{\emph{GAIL}}}
\newcommand\GAIfO{\mbox{\emph{GAIfO}}}
\newcommand\TCN{\mbox{\emph{TCN}}}
\newcommand\BCO{\mbox{\emph{BCO}}}
\newcommand\GAN{\mbox{\emph{GAN}}}
\newcommand\MDP{\mbox{\emph{MDP}}}
\newcommand\TRPO{\mbox{\emph{TRPO}}}
\newcommand\PPO{\mbox{\emph{PPO}}}
\newcommand\ILPO{\mbox{\emph{ILPO}}}

\usepackage{fancyhdr}

\title{Imitation Learning from Video by Leveraging Proprioception}

\author{
Faraz Torabi$^1$
\and
Garrett Warnell$^2$\And
Peter Stone$^1$
\affiliations
$^1$The University of Texas at Austin\\
$^2$Army Research Laboratory\\
\emails
\{faraztrb, pstone\}@cs.utexas.edu,
garrett.a.warnell.civ@mail.mil
}

\begin{document}
\thispagestyle{fancy}
\maketitle

\begin{abstract}
Classically, imitation learning algorithms have been developed for idealized situations, e.g., the demonstrations are often required to be collected in the exact same environment and usually include the demonstrator's actions.
Recently, however, the research community has begun to address some of these shortcomings by offering algorithmic solutions that enable imitation learning from observation (\IfO), e.g., learning to perform a task from \emph{visual}
demonstrations that may be in a different environment and do not include
actions.
Motivated by the fact that agents often also have access to their own internal states (i.e., \emph{proprioception}), we propose and study an \IfO ~algorithm that leverages this information in the policy learning process.
The proposed architecture learns policies over proprioceptive state representations and compares the resulting trajectories visually to the demonstration data.
We experimentally test the proposed technique on several MuJoCo domains and show that it outperforms other imitation from observation algorithms by a large margin.
\end{abstract}

\section{Introduction}

Imitation learning \cite{schaal1997learning,argall2009survey,osa2018algorithmic} is a popular method by which artificial agents learn to perform tasks.
In the imitation learning framework, an expert agent provides demonstrations of a task to a learning agent, and the learning agent attempts to mimic the expert.
Unfortunately, many existing imitation learning algorithms have been designed for idealized situations, e.g., they require that the demonstrations be collected in the exact same environment as the one that the imitator is in and/or that the demonstrations include the demonstrator's actions, i.e., the internal control signals that were used to drive the behavior.
These limitations result in the exclusion of a large amount of existing resources, including a large number of videos uploaded to the internet.
For example, 300 hours of video are uploaded to YouTube every minute\footnote{\url{https://bit.ly/2quPG6O}}, many of which include different types of tasks being performed.
Without new imitation learning techniques, none of this video can be used to instruct artificial agents.

Fortunately, the research community has recently begun to focus on addressing the above limitations by considering the specific problem of imitation from observation (\IfO) \cite{liu2017imitation,torabi2019recent}.
\IfO ~considers situations in which agents attempt to learn tasks by observing demonstrations that contain only state information (e.g., videos).
Among \IfO ~algorithms that learn tasks by watching videos, most attempt to learn imitation policies that rely solely on self-observation through video, i.e., they use a convolutional neural network (CNN) that maps images of themselves to actions.
However, in many cases, the imitating agent also has access to its own \emph{proprioceptive} state information, i.e., direct knowledge of itself such as the joint angles and torques associate with limbs.
In this paper, we argue that \IfO ~algorithms that ignore this information are missing an opportunity that could potentially improve the performance and the efficiency of the learning process.
Therefore, we are interested here in \IfO ~algorithms that can make use of both visual and proprioceptive state information.

In this paper, we build upon our previous work \cite{torabi2018generative} proposing an algorithm that uses a \GAN-like \cite{goodfellow2014generative} architecture to learn tasks perform \IfO ~directly from videos.
Unlike our prior work, however, our method \emph{also} uses proprioceptive information from the imitating agent during the learning process.
We hypothesize that the addition of such information will improve both learning speed and the final performance of the imitator, and we test this hypothesis experimentally in several standard simulation domains.
We compare our method with other, state-of-the-art approaches that do not leverage proprioception, and our results validate our hypothesis, i.e., the proposed technique outperforms the others by a large margin.

The rest of this paper is organized as follows.
In Section \ref{sec:related-work}, we review related work in imitation from observation.
In Section \ref{sec:background}, we review technical details surrounding Markov decision processes, imitation learning, and \IfO.
The proposed algorithm is presented in Section \ref{sec:algorithm}, and we describe the experiments that we have performed in Section \ref{sec:experiments}.

\section{Related Work}
\label{sec:related-work}
In this section, we review research in imitation learning, plan/goal recognition by mirroring, and recent advances in imitation from observation (\IfO).

Conventionally, imitation learning is used in autonomous agents to learn tasks from demonstrated {\em state-action} trajectories. The algorithms developed for this task can be divided into two general categories, (1) behavioral cloning \cite{bain1999framework,ross2011reduction,daftry2016learning} in which the agents learn a direct mapping from the demonstrated states to the actions, and (2) inverse reinforcement learning (IRL) \cite{abbeel2004apprenticeship,bagnell2007boosting,baker2009action} in which the agents first learn a reward function based on the demonstrations and then learn to perform the task using a reinforcement learning (\RL) ~\cite{sutton1998reinforcement} algorithm.

In contrast, imitation from observation (\IfO) is a framework for learning a task from {\em state-only} demonstrations. This framework has recently received a great deal of attention from the research community.
The \IfO ~algorithms that have been developed can be categorized as either (1) model-based, or (2) model-free.
Model-based algorithms require the agent to learn an explicit model of its environment as part of the imitation learning process.
One algorithm of this type is behavioral cloning from observation (\BCO) \cite{torabi2018behavioral}, in which the imitator learns a dynamics model of its environment using experience collected by a known policy, and then uses this model to infer the missing demonstrator actions.
Using the inferred actions, the imitator then computes an imitation policy using behavioral cloning \cite{bain1995a}. Another approach of this type is reinforced inverse dynamics modeling (RIDM) \cite{torabi2019RIDM} which also learns a model of its environment using an exploration policy and then it further optimizes the model using a sparse reward function. It is shown that in some experiments the algorithm can even outperform the expert.
Another model-based approach to \IfO ~is imitating latent policies from observation (\ILPO) \cite{edwards2018imitating}.
Given the current state of the expert, this approach predicts the next state using a latent policy and a forward dynamics model.
It then uses the difference between the predicted state and the actual demonstrator next state to update both the model and the imitation policy.
Afterwards, the imitator interacts with its environment to correct the action labels.

Model-free algorithms, on the other hand, do not require any sort of model to learn imitation policies.
One set of approaches of this type learns a time-dependent representation of tasks and then relies on hand-designed, time-aligned reward functions to learn the task via \RL.
For example, \citeauthor{sermanet2017time} \shortcite{sermanet2017time} propose an algorithm that learns an embedding function using a triplet loss function that seeks to push states that are close together in time closer together in the embedded space, while pushing other states further away.
\citeauthor{liu2017imitation} \shortcite{liu2017imitation} also propose a new architecture to learn a state representation---specifically, one that is capable of handling viewpoint differences.
\citeauthor{gupta2017learning} \shortcite{gupta2017learning} also propose a neural network architecture to try to learn a state representation that can overcome possible embodiment mismatch between the demonstrator and the imitator.
Each of these approaches requires multiple demonstrations of the same task to be time-aligned, which is typically not a realistic assumption.
\citeauthor{aytar2018playing} \shortcite{aytar2018playing} propose an \IfO ~algorithm that first learns an embedding using a self-supervised objective, and then constructs a reward function based on the embedding representation difference between the current state of the imitator and a specific checkpoint generated by the visual demonstration.
\citeauthor{goo2018learning} \shortcite{goo2018learning} propose an algorithm that uses a shuffle-and-learn style \cite{misra2016shuffle} loss in order to train a neural network that can predict progress in the task which can then be used as the reward function.

Another set of model-free algorithms follow a more end-to-end approach to learning policies directly from observations.
An algorithm of this type is generative adversarial imitation from observation (\GAIfO) \cite{torabi2018generative}, which uses a \GAN -like architecture to bring the state transition distribution of the imitator closer to that of the demonstrator.
Another approach of this type is the work of \citeauthor{merel2017learning} \shortcite{merel2017learning}, which is concerned instead with single state distributions.
\citeauthor{stadie2017third} \shortcite{stadie2017third} also propose an algorithm in this space that combines adversarial domain confusion methods \cite{ganin2016domain} with adversarial imitation learning algorithms in an attempt to overcome changes in viewpoint.
The method we propose in this paper also belongs to the category of end-to-end model-free imitation from observation algorithms.
However, it is different from the algorithms discussed above in that we explicitly incorporate the imitator's proprioceptive information in the learning process in order to study the improvement such information can make with respect to the performance and speed of the learning process.

A method that is closely related to imitation from observation is plan/goal recognition through mirroring \cite{vered2016online,vered2018towards} in that in it attempts to infer higher-level variables such as the goal or the future plan by observing other agents. However, in plan and goal recognition the observer already has fixed controllers, and then uses these controllers to match/explain the observed agent in order to infer their goal/plan. In imitation from observation, on the other hand, the agent seeks to learn a controller that the observer can use to imitate the observed agent.

%

\section{Background}
\label{sec:background}
In this section, we establish notation and provide background information about Markov decision processes (\MDP s) and adversarial imitation learning.

\subsection{Notation}
We consider artificial learning agents operating in the framework of Markov decision processes (\MDP s).
An \MDP ~can be described as a 6-tuple $\M = \{\SSS, \A, P, r, \gamma\}$, where $\SSS$ and $\A$ are state and action spaces, $P(s_{t+1}|s_t, a_t)$ is a function which represents the probability of an agent transitioning from state $s_t$ at time $t$ to $s_{t+1}$ at time $t+1$ by taking action $a_t$, $r:\SSS \times \A \rightarrow \R$ is a function that represents the reward feedback that the agent receives after taking a specific action at a given state, and $\gamma$ is a discount factor.
In the context of the notation established above, we are interested here in learning a policy $\pi:\SSS \rightarrow \A$ that can be used to select an action at each state.

In this paper, we refer to $s$ as the \emph{proprioceptive} state, i.e., $s$ is the most basic, internal state information available to the agent (e.g., the joint angles of a robotic arm).
Since we are also concerned with visual observations of agent behavior, we denote these observations as $o \in \OO$, i.e., an image of the agent at time $t$ is denoted as $o_t$.
The visual observations of the agent are determined both by the agent's current proprioceptive state $s$, and also other factors relating to image formation such as camera position.
Importantly, due to phenomena such as occlusion, it is not always possible to infer $s$ from $o$ alone.

In imitation learning (\IL), agents do not receive reward feedback $r$.
Instead, they have access to expert demonstrations of the task.
These demonstrations $\tau_e = \{(s_t,a_t)\}$ are composed of the state and action sequences experienced by the demonstrator.
Here, however, we specifically consider the problem of imitation from observation (\IfO), in which the agent only has access to sequences of visual observations of the demonstrator performing the task, i.e., $\tau_e = \{o_t\}$.

\subsection{Adversarial Imitation Learning}
Generative adversarial imitation learning (\GAIL) is a recent imitation learning algorithm developed by \citeauthor{ho2016generative} \shortcite{ho2016generative} that formulates the problem of finding an imitating policy as that of solving the following optimization problem:
\begin{equation}\label{gail}
\begin{split}
\min_{\pi \in \prod} \displaystyle{\max_{D \in (0,1)^{\mathcal{S} \times \mathcal{A}}}} & -\lambda_H H(\pi) + \mathbb{E}_\pi[\log(D(s,a)] +\\ &\mathbb{E}_{\pi_E}[\log(1-D(s,a))]\;,
\end{split}
\end{equation}
where $H$ is the entropy function, and the discriminator function $D:\mathcal{S} \times \mathcal{A} \rightarrow (0,1)$ can be thought of as a classifier trained to differentiate between the state-action pairs provided by the demonstrator and those experienced by the imitator.
The objective in (\ref{gail}) is similar the one used in generative adversarial networks (\GAN s) \cite{goodfellow2014generative}, and the associated algorithm can be thought of as trying to induce an imitator state-action occupancy measure that is similar to that of the demonstrator.
Even more recently, there has been research on methods that seek to improve on \GAIL ~by, e.g., increasing sample efficiency \cite{kostrikov2018discriminatoractorcritic,sasaki2018sample} and improving reward representation \cite{fu2018learning,qureshi2018adversarial}.

The method we propose in this paper is most related to generative adversarial imitation from observation \cite{torabi2018generative}, which models the imitating policy using a randomly-initialized convolutional neural network, executes the policy to generate recorded video of the imitator's behavior, and then trains a discriminator to differentiate between video of the demonstrator and video of the imitator.
Next, it uses the discriminator as a reward function for the imitating agent (higher rewards corresponding to behavior the discriminator classifies as coming from the demonstrator), and uses a policy gradient technique (e.g., \TRPO ~\cite{schulman2015trust}) to update the policy.
The process repeats until convergence.
This algorithm differs from what we propose in that \GAIfO ~uses visual data {\it both} in the process of discriminator {\it and} policy learning.
That is, the learned behavior policy maps images $o$ to actions using a convolutional neural network.
The technique we propose, on the other hand, leverages proprioceptive information in the policy learning step, instead learning policies that map proprioceptive states $s$ to actions using a multilayer perceptron architecture.

\begin{figure*}[!ht]
\centering
\begin{tikzpicture}

	\draw[use as bounding box, transparent] (-3,0.3) rectangle (14, 3.3);
	\node [align=center] at (-1.8, 2) {Proprioceptive\\ features ($s_t$)\\
	(e.g. joint angles)};
	\node[] (input image) at (-0.2,2.4) {\includegraphics[height=30mm]{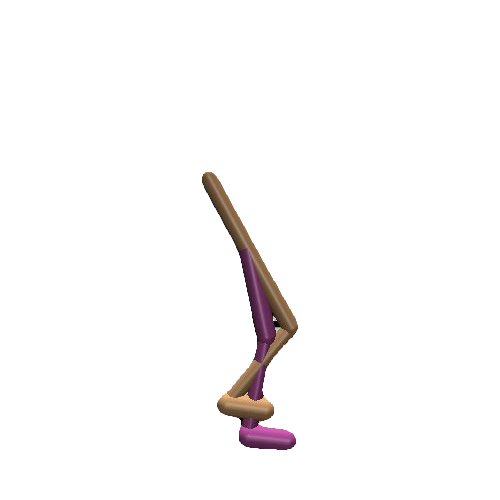}};
	\draw[->, line width=0.2mm] (.3,2) -- (1.1,2);
	\draw[fill=brown,draw=black] (1.1,1) -- (1.1,3) -- (2.1,3) -- (2.1,1) -- (1.1,1);
	\node [align=center] at (1.6, 2) {MLP\\ ($\pi_\theta$)};
	\draw[-, line width=0.2mm] (2.1,2) -- (2.7,2);
	\draw[fill=light-blue,draw=black] (2.9,1.3) -- (2.9,2.7) -- (3.1,2.7) -- (3.1,1.3) -- (2.9,1.3);
	\draw[-, line width=0.2mm] (2.9,2.5) -- (3.1,2.5);
	\draw[-, line width=0.2mm] (2.9,1.5) -- (3.1,1.5);
	\draw[-, line width=0.2mm] (2.9,1.7) -- (3.1,1.7);
	\node [align=center] at (3, 1.9) {.};
	\node [align=center] at (3, 2.1) {.};
	\node [align=center] at (3, 2.3) {.};
	\node [align=center] at (3, 1) {$a_t$};
	\draw[->, line width=0.2mm] (3.3,2) -- (3.9,2);
	\draw[fill=light-yellow,draw=black] (3.9,1.5) -- (3.9,2.5) -- (4.7,2.5) -- (4.7,1.5) -- (3.9,1.5);
	\node [align=center] at (4.3, 2) {Env};
	\draw[-, line width=0.2mm] (4.7,2) -- (5.5,2);
	\node[] (input image) at (6.35,2.2) {\includegraphics[height=15mm,width=15mm]{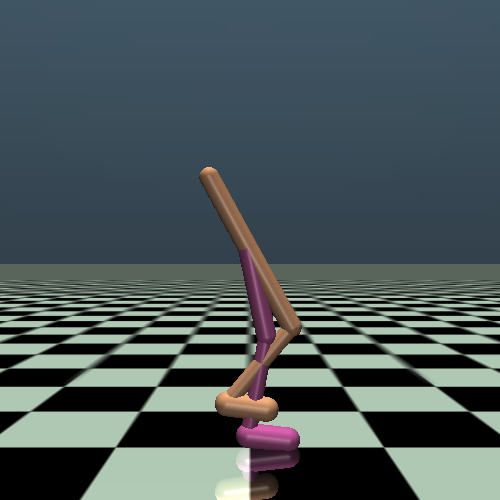}};
	\draw[-, line width=0.2mm] (5.6,2.95) -- (5.8,2.75);
	\draw[-, line width=0.2mm] (7.1,2.95) -- (7.3,2.75);
	\draw[-, line width=0.2mm] (5.6,1.45) -- (5.8,1.25);
	\node[] (input image) at (6.55,2) {\includegraphics[height=15mm,width=15mm]{walker2.png}};
	\node [align=center] at (6.45, .8) {$\tau_i=\{o_{t-2}:o_{t+1}\}$};
	\draw[->, line width=0.2mm] (7.5,2) -- (7.9,2) -- (7.9,2.7) -- (8.9,2.7) -- (8.9,2.5);
	
	\draw[fill=brown,draw=black] (8.5,2.5) -- (10.5,2.5) -- (10,1.5) -- (9,1.5) -- (8.5,2.5);
	\node [align=center] at (9.5, 2) {CNN\\ ($D_\phi$)};
	\draw[->, line width=0.2mm] (9.5,1.5) -- (9.5,1);
	\node [align=center] at (9.5, 0.75) {v};
	
	\node[] (input image) at (12.4,2.2) {\includegraphics[height=15mm,width=15mm]{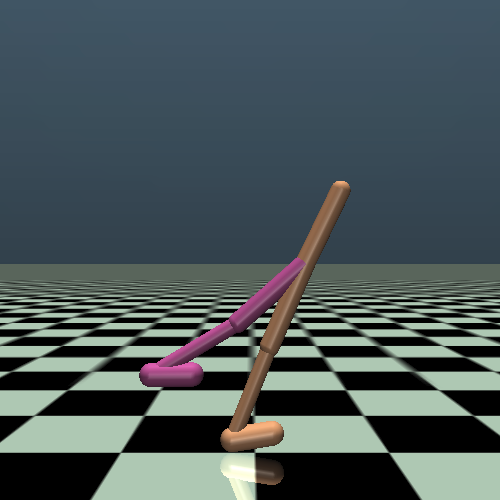}};
	\draw[-, line width=0.2mm] (11.65,2.95) -- (11.85,2.75);
	\draw[-, line width=0.2mm] (13.15,2.95) -- (13.35,2.75);
	\draw[-, line width=0.2mm] (11.65,1.45) -- (11.85,1.25);
	\node[] (input image) at (12.6,2) {\includegraphics[height=15mm,width=15mm]{walker.png}};
	\node [align=center] at (12.5, .8) {$\tau_e=\{o_{t-2}:o_{t+1}\}$};
	\draw[->, line width=0.2mm] (11.5,2) -- (11.1,2) -- (11.1,2.7) -- (10.1,2.7) -- (10.1,2.5);
	
%
	
\end{tikzpicture}
\caption{A diagrammatic representation of our algorithm. A multilayer perceptron (MLP) is used to model the policy, which takes the proprioceptive features $s_t$ as the input and outputs an action $a$. The agent then executes the action in its environment. While the agent executes the policy, a video of the resulting behavior is recorded. Stacks of four consecutive grayscale images $(o_{t-2}:o_{t+1})$ from both the demonstrator and the imitator are then prepared as the input for the discriminator, which is trained to discriminate between data coming from these two sources. Finally, the discriminator function is then used as the reward function to train the policy using \PPO ~(not shown).}
\label{fig:alg}
\end{figure*}

\section{Proposed Method}
\label{sec:algorithm}
As presented in Section \ref{sec:background}, we are interested in the problem of imitation from observation (\IfO), where an imitating agent has access to visual demonstrations, $\tau_e = \{o_t\}$, of an expert performing a task, and seeks to learn a behavior that is approximately the same as the expert's.
In many previous approaches to this problem, the imitator selects actions on the basis of visual self-observation alone (i.e., using images of itself).
We hypothesize that also leveraging available proprioceptive state information, $s$, during the learning process will result in better and faster learning.

Inspired by \GAIL , our algorithm is comprised of two pieces: (1) a generator, which corresponds to the imitation policy, and (2) a discriminator, which serves as the reward function for the imitator.
We model the imitation policy as a multilayer perceptron (MLP), $\pi_\theta$.
The imitating agent, being aware of its own proprioceptive features $s$, feeds them into the policy network and receives as output a distribution over actions from which the selected action $a$ can be sampled.
The imitator then executes this action and we record a video of the resulting behavior.
After several actions have been executed, we have accumulated a collection of visual observations of the imitator's behavior, $\tau_i = \{o\}$.

Meanwhile, we use a convolutional neural network as a discriminator $D_\phi$.
Given visual observations of the demonstrator, $\tau_e$, and observations of the imitator, $\tau_i$, we train the discriminator to differentiate between the data coming from these different sources.
Since single video frames lack observability in most cases, we instead stack four frames, $\{o_{t-2}, o_{t-1}, o_t, o_{t+1}\}$, and feed this stack as input to the discriminator.

\begin{algorithm}[t]
	\caption{ }\label{alg}
	\begin{algorithmic}[1]
		\STATE Initialize policy $\pi_\theta$ randomly 
		\STATE Initialize discriminator $D_\phi$ randomly
		\STATE Obtain visual demonstrations $\tau_e=\{o\}$
		\FOR {$i \leftarrow 0$ \textit{\textbf{to}} $N$}
			\STATE Execute $\pi_\theta$ and record video observation $\tau_i=\{o\}$
			\STATE Update the discriminator $D_\phi$ using loss 
			\begin{equation*}\label{eq:disc}
			\begin{split}
				- & \Big(\mathbb{E}_{\tau_i} [\log(D_\phi(o_{t-2}:o_{t+1}))]+ \\& \mathbb{E}_{\tau_e} [\log(1-D_\phi(o_{t-2}:o_{t+1}))]\Big)
			\end{split}
			\end{equation*}
			\STATE Update $\pi_\theta$ by performing \mbox{\emph{PPO}} updates with gradient steps of
			
			\begin{equation*}\label{eq:step}
			\begin{split}
				\mathbb{E}_{\tau_i} [\nabla_\theta \log \pi_\theta(a|s) Q(s,a)] - \lambda \nabla_\theta H(\pi_\theta),
			\end{split}
			\end{equation*}
			where
			\begin{equation*}\label{eq:step}
			\begin{split}
				& Q(\hat{s}_t,\hat{a}_t) = \\& -\mathbb{E}_{\tau_i} [\log(D_\phi(o_{t-2}:o_{t+1}))| s_0=\hat{s}_t, a_0=\hat{a}_t]
			\end{split}
			\end{equation*}
		\ENDFOR
	\end{algorithmic}
\end{algorithm}

We train the discriminator to output values closer to zero for the transitions coming from the expert, and values closer to one for those coming from the imitator.
Therefore, the discriminator aims to solve the following optimization problem:
\begin{equation}\label{eq:disc}
\begin{split}
\displaystyle{\max_\phi}& \Big(\mathbb{E}_{\tau_i} [\log(D_\phi(o_{t-2}:o_{t+1}))]+\\& \mathbb{E}_{\tau_e} [\log(1-D_\phi(o_{t-2}:o_{t+1}))]\Big) \; .
\end{split}
\end{equation}
The lower the value outputted by the discriminator, the higher the chance of the input being from the expert.
Recall that the objective for the imitator is to mimic the demonstrator, which can be thought of as fooling the discriminator.
Therefore, we use
\begin{equation}\label{eq:rew}
\begin{split}
-\Big(\mathbb{E}_{\tau_i} [\log(D_\phi(o_{t-2}:o_{t+1}))]\Big)\\[5pt]
\end{split}
\end{equation}
as the reward to update the imitation policy using \RL .
In particular, we use proximal policy optimization (\PPO) \cite{schulman2017proximal} with gradient steps of
\begin{equation}\label{eq:step}
\begin{split}
\mathbb{E}_{\tau_i} [\nabla_\theta \log \pi_\theta(a|s) Q(s,a)] - \lambda \nabla_\theta H(\pi_\theta),\\[5pt]
\end{split}
\end{equation}
where $Q(s,a)$ is the state-action value, i.e. the potential reward that the agent receives starting from $s$ and taking action $a$:
\begin{equation}\label{eq:step}
\begin{split}
& Q(\hat{s}_t,\hat{a}_t) = \\&-\mathbb{E}_{\tau_i} [\log(D_\phi(o_{t-2}:o_{t+1}))| s_0=\hat{s}_t, a_0=\hat{a}_t].\\[5pt]
\end{split}
\end{equation}

As presented, our algorithm uses the visual information in order to learn the reward function by comparing visual data generated by the imitator and the demonstrator.
It also takes advantage of proprioceptive state features in the process of policy learning by learning a mapping from those features to actions using a reinforcement learning algorithm.
Pseudocode and a diagrammatic representation of our proposed algorithm are presented in Algorithm \ref{alg} and Figure \ref{fig:alg}, respectively. 


\section{Experiments}
\label{sec:experiments}
The algorithm introduced above combines proprioceptive state information with video observations in an adversarial imitation learning paradigm.
We hypothesize that using the extra state information in the proposed way will lead to both faster imitation learning and better performance on the imitated task when compared to similar techniques that ignore proprioception.
In this section, we describe the experimental procedure by which we evaluated this hypothesis, and discuss the results.

\subsection{Setup}
We evaluated our method on a subset of the continuous control tasks available via OpenAI Gym \cite{1606.01540} and the MuJoCo simulator \cite{todorov2012mujoco}: MountainCarContinuous, InvertedPendulum, InvertedDoublePendulum, Hopper, Walker2d, HalfCheetah.

To generate the demonstration data, we first trained an expert agents using pure reinforcement learning (i.e., not from imitation).
More specifically, we used proximal policy optimization (\PPO) \cite{schulman2017proximal} and the ground truth reward function provided by OpenAI Gym.
After the expert agents were trained, we recorded $64 \times 64$, $30$-fps video demonstrations of their behavior. 

We compared the proposed method with three other imitation from observation algorithms that do \emph{not} exploit the imitator's proprioceptive state information: Time Contrastive Networks (TCN) \cite{sermanet2017time}, Behavioral Cloning from Observation (BCO) \cite{torabi2018behavioral}, Generative Adversarial Imitation Learning (GAIfO) \cite{torabi2018generative,torabi2019adversarial}\footnote{The considered domains, methods, and implementations are presented in more detail in the longer version of the paper on arXiv \cite{torabi2019imitation}}.

\subsection{Results}
We hypothesized that our method would outperform the baselines with respect to two criteria: (1) the final performance of the trained imitator, i.e., how the imitator performs the task compared to the demonstrator (as measured by the ground truth reward functions), and (2) the speed of the imitation learning process as measured by number of learning iterations.
The results shown here were generated using ten independent trials, where each trial used a different random seed to initialize the environments, model parameters, etc.

\begin{figure}[!ht]
	\centering
	\includegraphics[scale=1.5,width=\linewidth]{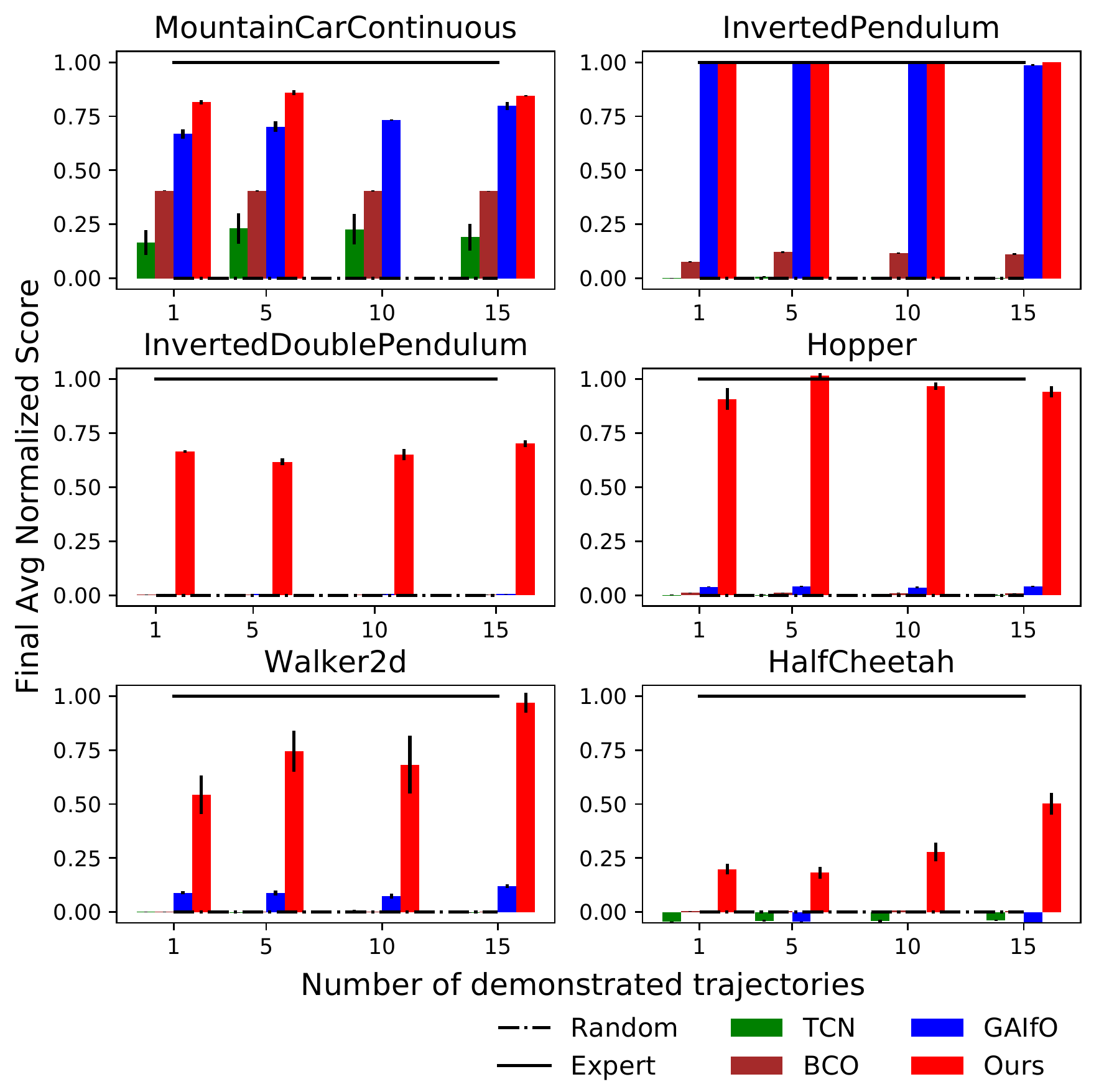}
	\caption{The rectangular bars and error bars represent the mean normalized return and the standard error, respectively, as measured over 1000 trials. The normalized values have been scaled in such a way that expert and random performance are $1.0$ and $0.0$, respectively. The x-axis represents the number of available video demonstration trajectories.}
	\label{fig:performance}
\end{figure}

Figure \ref{fig:performance} depicts our experimental results pertaining to the first criterion, i.e., the final task performance of trained imitating agents in each domain.
The rectangular bars and error bars represent the mean return and the standard error, respectively, as measured over $1000$ trajectories.
We report performance using a normalized task score, i.e., scores are scaled in such a way that the demonstrating agent's performance corresponds to $1.0$ and the performance of an agent with random behavior corresponds to $0.0$.
The x-axis represents the number of demonstration trajectories, i.e., videos, available to the imitator.
In general, it can be seen that the proposed method indeed outperforms the baselines in almost all cases, which shows that using the available proprioceptive state information can make a remarkable difference in the final task performance achieved by imitation learning.
In the particular case of InvertedPendulum, both \GAIfO ~and the proposed method achieve a final task performance equal to that of the demonstrator, likely due to the simplicity of the task.
However, for the rest of the tasks, it can be clearly seen that the proposed approach performs better than \GAIfO \footnote{Note that the performance of \GAIfO ~on Hopper is different from what was presented in the \GAIfO ~paper \cite{torabi2018generative}. We hypothesize that the reason is twofold: (1) different physics engines---MuJoCo is used in this paper, but in the previous work \cite{torabi2018generative} Pybullet \cite{coumans2016pybullet} was used , and (2) differences in video appearance---in this work we do not alter the default simulator parameters, whereas in the previous work \cite{torabi2018generative} some of the parameters were modified such as the colors used in the video frames in order to increase the contrast between the agent and the background.}.
Further, we can see that increasing the number of demonstrated trajectories results in increased task performance.

\begin{figure}[!ht]
	\centering
	\includegraphics[scale=.48]{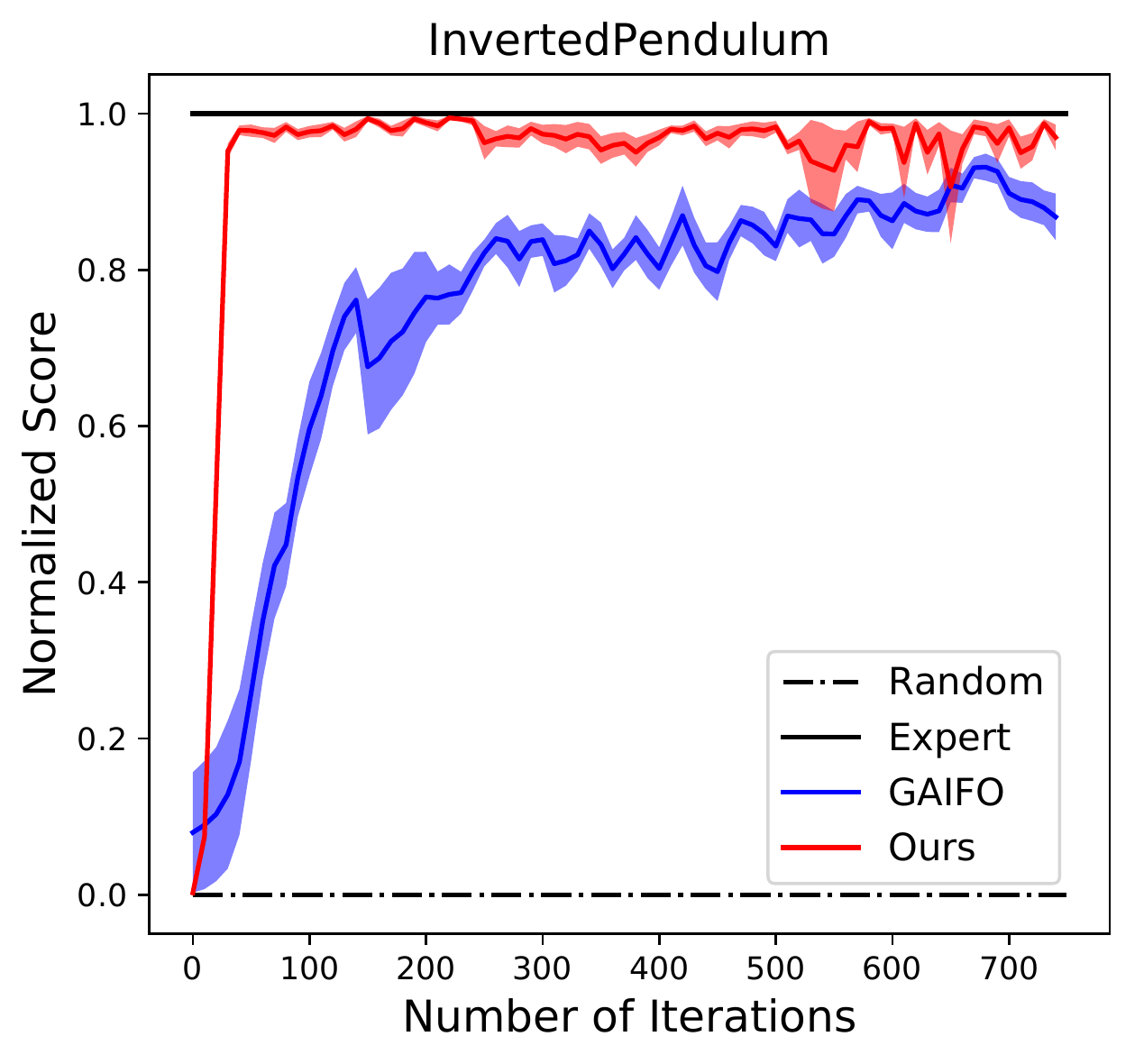}
	\caption{Performance of imitation agents with respect to the number of iterations (N). Solid colored lines represent the mean return and shaded areas represent standard errors. The returns are scaled so that the performance of the expert and random policies be zero and one, respectively.}
	\label{fig:process}
\end{figure}

To validate our hypothesis with respect to learning speed, we also studied the transient performance of the various learning algorithms.
Because only one other method, \GAIfO , performed as well as the expert in only one domain, InvertedPendulum,  Figure \ref{fig:process} only depicts the results for these algorithms in that domain.
The x-axis shows the number of iterations, i.e., the number of update cycles for both the policy and the discriminator.
Since updating the policy requires interaction with the environment, a smaller number of iterations also corresponds to less overhead during the learning process.
As shown in the figure, our method converges to expert-level performance much faster than \GAIfO, which supports our hypothesis that leveraging proprioception speeds the imitation learning process.

In Figure \ref{fig:performance}, we can see that two of the baseline methods---\BCO ~and \TCN---do not achieve task performance anywhere near that of the expert.

For InvertedPendulum and InvertedDoublePendulum, we suspect that \TCN ~performs poorly due to possible overfitting of the learned state embedding to the specific demonstrations and, therefore, does not generalize well toward supporting the overall goal of keeping the pendulum balanced above the rod.
For Hopper, Walker2d, and HalfCheetah, the poor performance of \TCN ~may be due to the fact that the tasks are cyclical in nature and therefore not well-suited to the time-dependent learned state embedding. \TCN ~performs relatively better in MountainCarContinuous, compared to other domains because this domain does have the properties required by \TCN.
As for \BCO , we posit that the low performance is due to the well-known compounding-error issue present in behavioral cloning.

One interesting thing to note is that Walker2d results in larger error bars for our technique than those seen for any of the other domains.
We hypothesize that the reason for this is that the video frames provide very poor information regarding the state of the demonstrator---here, the agent has two legs, which sometimes results in occlusion and, therefore, uncertainty regarding which action the agent should take.

Finally, we can see that the proposed technique performs the most poorly in the HalfCheetah domain.
We hypothesize that this is due to the speed at which the demonstrator acts: frame-to-frame differences are large, e.g., three to four consecutive frames cover a complete cycle of the agent jumping forwards.
This rate of change may make it difficult for our discriminator to extract a pattern of behavior, which, consequently, would make it much more difficult for the agent to move its behavior closer to that of the demonstrator.
Therefore, one way that performance might be improved is to increase the frame rate at which the demonstrations are sampled.
Another way, as suggested by Figure \ref{fig:performance}, would be to increase the number of demonstration trajectories beyond what is shown here.

\section{Conclusion and Future Work}
In this paper, we hypothesized that including proprioception would be beneficial to the learning process in the \IfO ~paradigm.
To test this hypothesis, we presented a new imitation from observation algorithm that leverages both available visual and proprioceptive information.
It uses visual information to compare the imitator's behavior to that of the demonstrator, and uses this comparison as a reward function for training a policy over proprioceptive states.
We showed that leveraging this state information can significantly improve both the performance and the efficiency of the learning process. 

However, to achieve the end-goal of true imitation from observation, several challenges remain. For example, \IfO ~algorithms should be able to overcome embodiment mismatch (the imitator and the demonstrator have different embodiments), and viewpoint mismatch (the visual demonstrations are recorded from different viewpoints.). Resolving these limitations is a natural next step for extending this research. Another way to improve upon the proposed method is to attempt to make the training more reliable by incorporating techniques developed to improve the stability of \GAN s, such as the work of \citeauthor{arjovsky2017wasserstein} \shortcite{arjovsky2017wasserstein}.
Further, to the best of our knowledge, nobody has been able to deploy \GAN-like methods on real robots due to high sample complexity.
Therefore, techniques that seek to improve the learning process with respect to this metric should also be investigated further.

\section*{Acknowledgments}
This work has taken place in the Learning Agents Research
Group (LARG) at the Artificial Intelligence Laboratory, The University
of Texas at Austin.  LARG research is supported in part by grants from
the National Science Foundation (IIS-1637736, IIS-1651089,
IIS-1724157), the Office of Naval Research (N00014-18-2243), Future of
Life Institute (RFP2-000), Army Research Lab, DARPA, Intel, Raytheon,
and Lockheed Martin.  Peter Stone serves on the Board of Directors of
Cogitai, Inc.  The terms of this arrangement have been reviewed and
approved by the University of Texas at Austin in accordance with its
policy on objectivity in research.
\bibliographystyle{named}
\bibliography{ijcai19}

\begin{thebibliography}{}

\bibitem[\protect\citeauthoryear{Abbeel and
  Ng}{2004}]{abbeel2004apprenticeship}
Pieter Abbeel and Andrew~Y Ng.
\newblock Apprenticeship learning via inverse reinforcement learning.
\newblock In {\em Proceedings of the twenty-first International Conference on
  Machine learning}, page~1. ACM, 2004.

\bibitem[\protect\citeauthoryear{Argall \bgroup \em et al.\egroup
  }{2009}]{argall2009survey}
Brenna~D Argall, Sonia Chernova, Manuela Veloso, and Brett Browning.
\newblock A survey of robot learning from demonstration.
\newblock {\em Robotics and Autonomous Systems}, 57(5):469--483, 2009.

\bibitem[\protect\citeauthoryear{Arjovsky \bgroup \em et al.\egroup
  }{2017}]{arjovsky2017wasserstein}
Martin Arjovsky, Soumith Chintala, and L{\'e}on Bottou.
\newblock Wasserstein generative adversarial networks.
\newblock In {\em ICML}, pages 214--223, 2017.

\bibitem[\protect\citeauthoryear{Aytar \bgroup \em et al.\egroup
  }{2018}]{aytar2018playing}
Yusuf Aytar, Tobias Pfaff, David Budden, Thomas Paine, Ziyu Wang, and Nando
  de~Freitas.
\newblock Playing hard exploration games by watching youtube.
\newblock In {\em NeurIPS}, pages 2935--2945, 2018.

\bibitem[\protect\citeauthoryear{Bagnell \bgroup \em et al.\egroup
  }{2007}]{bagnell2007boosting}
JA~Bagnell, Joel Chestnutt, David~M Bradley, and Nathan~D Ratliff.
\newblock Boosting structured prediction for imitation learning.
\newblock In {\em Advances in Neural Information Processing Systems}, pages
  1153--1160, 2007.

\bibitem[\protect\citeauthoryear{Bain and Sammut}{1995}]{bain1995a}
Michael Bain and Claude Sammut.
\newblock A framework for behavioral cloning.
\newblock {\em Machine Intelligence 14}, 1995.

\bibitem[\protect\citeauthoryear{Bain and Sommut}{1999}]{bain1999framework}
Michael Bain and Claude Sommut.
\newblock A framework for behavioural claning.
\newblock {\em Machine Intelligence 15}, 15:103, 1999.

\bibitem[\protect\citeauthoryear{Baker \bgroup \em et al.\egroup
  }{2009}]{baker2009action}
Chris~L Baker, Rebecca Saxe, and Joshua~B Tenenbaum.
\newblock Action understanding as inverse planning.
\newblock {\em Cognition}, 113(3):329--349, 2009.

\bibitem[\protect\citeauthoryear{Brockman \bgroup \em et al.\egroup
  }{2016}]{1606.01540}
Greg Brockman, Vicki Cheung, Ludwig Pettersson, Jonas Schneider, John Schulman,
  Jie Tang, and Wojciech Zaremba.
\newblock Open{AI} {G}ym, 2016.

\bibitem[\protect\citeauthoryear{Coumans and Bai}{2016
  2017}]{coumans2016pybullet}
Erwin Coumans and Yunfei Bai.
\newblock Pybullet, a python module for physics simulation for games, robotics
  and machine learning.
\newblock 2016-2017.

\bibitem[\protect\citeauthoryear{Daftry \bgroup \em et al.\egroup
  }{2016}]{daftry2016learning}
Shreyansh Daftry, J~Andrew Bagnell, and Martial Hebert.
\newblock Learning transferable policies for monocular reactive mav control.
\newblock In {\em International Symposium on Experimental Robotics}, pages
  3--11. Springer, 2016.

\bibitem[\protect\citeauthoryear{Edwards \bgroup \em et al.\egroup
  }{2019}]{edwards2018imitating}
Ashley~D Edwards, Himanshu Sahni, Yannick Schroeker, and Charles~L Isbell.
\newblock Imitating latent policies from observation.
\newblock {\em ICML}, 2019.

\bibitem[\protect\citeauthoryear{Fu \bgroup \em et al.\egroup
  }{2018}]{fu2018learning}
Justin Fu, Katie Luo, and Sergey Levine.
\newblock Learning robust rewards with adverserial inverse reinforcement
  learning.
\newblock In {\em ICLR}, 2018.

\bibitem[\protect\citeauthoryear{Ganin \bgroup \em et al.\egroup
  }{2016}]{ganin2016domain}
Yaroslav Ganin, Evgeniya Ustinova, Hana Ajakan, Pascal Germain, Hugo
  Larochelle, Fran{\c{c}}ois Laviolette, Mario Marchand, and Victor Lempitsky.
\newblock Domain-adversarial training of neural networks.
\newblock {\em JMLR}, 17(1):2096--2030, 2016.

\bibitem[\protect\citeauthoryear{Goo and Niekum}{2019}]{goo2018learning}
Wonjoon Goo and Scott Niekum.
\newblock One-shot learning of multi-step tasks from observation via activity
  localization in auxiliary video.
\newblock {\em ICRA}, 2019.

\bibitem[\protect\citeauthoryear{Goodfellow \bgroup \em et al.\egroup
  }{2014}]{goodfellow2014generative}
Ian Goodfellow, Jean Pouget-Abadie, Mehdi Mirza, Bing Xu, David Warde-Farley,
  Sherjil Ozair, Aaron Courville, and Yoshua Bengio.
\newblock Generative adversarial nets.
\newblock In {\em Advances in Neural Information Processing Systems}, 2014.

\bibitem[\protect\citeauthoryear{Gupta \bgroup \em et al.\egroup
  }{2017}]{gupta2017learning}
Abhishek Gupta, Coline Devin, YuXuan Liu, Pieter Abbeel, and Sergey Levine.
\newblock Learning invariant feature spaces to transfer skills with
  reinforcement learning.
\newblock In {\em International Conference on Learning Representations}, 2017.

\bibitem[\protect\citeauthoryear{Ho and Ermon}{2016}]{ho2016generative}
Jonathan Ho and Stefano Ermon.
\newblock Generative adversarial imitation learning.
\newblock In {\em Advances in Neural Information Processing Systems}, pages
  4565--4573, 2016.

\bibitem[\protect\citeauthoryear{Kostrikov \bgroup \em et al.\egroup
  }{2019}]{kostrikov2018discriminatoractorcritic}
Ilya Kostrikov, Kumar~Krishna Agrawal, Debidatta Dwibedi, Sergey Levine, and
  Jonathan Tompson.
\newblock Discriminator-actor-critic: Addressing sample inefficiency and reward
  bias in adversarial imitation learning.
\newblock In {\em ICLR}, 2019.

\bibitem[\protect\citeauthoryear{Liu \bgroup \em et al.\egroup
  }{2018}]{liu2017imitation}
YuXuan Liu, Abhishek Gupta, Pieter Abbeel, and Sergey Levine.
\newblock Imitation from observation: Learning to imitate behaviors from raw
  video via context translation.
\newblock 2018.

\bibitem[\protect\citeauthoryear{Merel \bgroup \em et al.\egroup
  }{2017}]{merel2017learning}
Josh Merel, Yuval Tassa, Sriram Srinivasan, Jay Lemmon, Ziyu Wang, Greg Wayne,
  and Nicolas Heess.
\newblock Learning human behaviors from motion capture by adversarial
  imitation.
\newblock {\em arXiv preprint arXiv:1707.02201}, 2017.

\bibitem[\protect\citeauthoryear{Misra \bgroup \em et al.\egroup
  }{2016}]{misra2016shuffle}
Ishan Misra, C~Lawrence Zitnick, and Martial Hebert.
\newblock Shuffle and learn: unsupervised learning using temporal order
  verification.
\newblock In {\em ECCV}, pages 527--544. Springer, 2016.

\bibitem[\protect\citeauthoryear{Osa \bgroup \em et al.\egroup
  }{2018}]{osa2018algorithmic}
Takayuki Osa, Joni Pajarinen, Gerhard Neumann, J~Andrew Bagnell, Pieter Abbeel,
  Jan Peters, et~al.
\newblock An algorithmic perspective on imitation learning.
\newblock {\em Foundations and Trends{\textregistered} in Robotics},
  7(1-2):1--179, 2018.

\bibitem[\protect\citeauthoryear{Pavse \bgroup \em et al.\egroup
  }{2019}]{torabi2019RIDM}
Brahma Pavse, Faraz Torabi, Josiah Hanna, Garrett Warnell, and Peter Stone.
\newblock Ridm: Reinforced inverse dynamics modeling for learning from a single
  observed demonstration.
\newblock In {\em International Conference on Machine Learning Workshop on
  Imitation, Intent, and Interaction (I3)}, 2019.

\bibitem[\protect\citeauthoryear{Qureshi \bgroup \em et al.\egroup
  }{2019}]{qureshi2018adversarial}
Ahmed~H. Qureshi, Byron Boots, and Michael~C. Yip.
\newblock Adversarial imitation via variational inverse reinforcement learning.
\newblock In {\em ICLR}, 2019.

\bibitem[\protect\citeauthoryear{Ross \bgroup \em et al.\egroup
  }{2011}]{ross2011reduction}
St{\'e}phane Ross, Geoffrey~J Gordon, and Drew Bagnell.
\newblock A reduction of imitation learning and structured prediction to
  no-regret online learning.
\newblock In {\em International Conference on Artificial Intelligence and
  Statistics}, pages 627--635, 2011.

\bibitem[\protect\citeauthoryear{Sasaki \bgroup \em et al.\egroup
  }{2019}]{sasaki2018sample}
Fumihiro Sasaki, Tetsuya Yohira, and Atsuo Kawaguchi.
\newblock Sample efficient imitation learning for continuous control.
\newblock In {\em ICLR}, 2019.

\bibitem[\protect\citeauthoryear{Schaal}{1997}]{schaal1997learning}
Stefan Schaal.
\newblock Learning from demonstration.
\newblock In {\em Advances in Neural Information Processing Systems}, pages
  1040--1046, 1997.

\bibitem[\protect\citeauthoryear{Schulman \bgroup \em et al.\egroup
  }{2015}]{schulman2015trust}
John Schulman, Sergey Levine, Pieter Abbeel, Michael Jordan, and Philipp
  Moritz.
\newblock Trust region policy optimization.
\newblock In {\em ICML}, pages 1889--1897, 2015.

\bibitem[\protect\citeauthoryear{Schulman \bgroup \em et al.\egroup
  }{2017}]{schulman2017proximal}
John Schulman, Filip Wolski, Prafulla Dhariwal, Alec Radford, and Oleg Klimov.
\newblock Proximal policy optimization algorithms.
\newblock {\em arXiv preprint arXiv:1707.06347}, 2017.

\bibitem[\protect\citeauthoryear{Sermanet \bgroup \em et al.\egroup
  }{2018}]{sermanet2017time}
Pierre Sermanet, Corey Lynch, Jasmine Hsu, and Sergey Levine.
\newblock Time-contrastive networks: Self-supervised learning from multi-view
  observation.
\newblock In {\em International Conference in Robotics and Automation (ICRA)},
  2018.

\bibitem[\protect\citeauthoryear{Stadie \bgroup \em et al.\egroup
  }{2017}]{stadie2017third}
Bradly~C Stadie, Pieter Abbeel, and Ilya Sutskever.
\newblock Third-person imitation learning.
\newblock In {\em ICLR}, 2017.

\bibitem[\protect\citeauthoryear{Sutton and
  Barto}{1998}]{sutton1998reinforcement}
Richard~S Sutton and Andrew~G Barto.
\newblock {\em Reinforcement learning: An introduction}, volume~1.
\newblock MIT press Cambridge, 1998.

\bibitem[\protect\citeauthoryear{Todorov \bgroup \em et al.\egroup
  }{2012}]{todorov2012mujoco}
Emanuel Todorov, Tom Erez, and Yuval Tassa.
\newblock Mujoco: A physics engine for model-based control.
\newblock In {\em IROS}, pages 5026--5033. IEEE, 2012.

\bibitem[\protect\citeauthoryear{Torabi \bgroup \em et al.\egroup
  }{2018}]{torabi2018behavioral}
Faraz Torabi, Garrett Warnell, and Peter Stone.
\newblock Behavioral cloning from observation.
\newblock In {\em IJCAI}, pages 4950--4957, 2018.

\bibitem[\protect\citeauthoryear{Torabi \bgroup \em et al.\egroup
  }{2019a}]{torabi2019adversarial}
Faraz Torabi, Garrett Warnell, and Peter Stone.
\newblock Adversarial imitation learning from state-only demonstrations.
\newblock In {\em AAMAS}, 2019.

\bibitem[\protect\citeauthoryear{Torabi \bgroup \em et al.\egroup
  }{2019b}]{torabi2018generative}
Faraz Torabi, Garrett Warnell, and Peter Stone.
\newblock Generative adversarial imitation from observation.
\newblock In {\em International Conference on Machine Learning Workshop on
  Imitation, Intent, and Interaction (I3)}, 2019.

\bibitem[\protect\citeauthoryear{Torabi \bgroup \em et al.\egroup
  }{2019c}]{torabi2019imitation}
Faraz Torabi, Garrett Warnell, and Peter Stone.
\newblock Imitation learning from video by leveraging proprioception.
\newblock {\em arXiv preprint arXiv:1905.09335}, 2019.

\bibitem[\protect\citeauthoryear{Torabi \bgroup \em et al.\egroup
  }{2019d}]{torabi2019recent}
Faraz Torabi, Garrett Warnell, and Peter Stone.
\newblock Recent advances in imitation learning from observation.
\newblock In {\em IJCAI}. AAAI Press, 2019.

\bibitem[\protect\citeauthoryear{Vered \bgroup \em et al.\egroup
  }{2016}]{vered2016online}
Mor Vered, Gal~A Kaminka, and Sivan Biham.
\newblock Online goal recognition through mirroring: Humans and agents.
\newblock In {\em The Fourth Annual Conference on Advances in Cognitive
  Systems}, 2016.

\bibitem[\protect\citeauthoryear{Vered \bgroup \em et al.\egroup
  }{2018}]{vered2018towards}
Mor Vered, Ramon~Fraga Pereira, Maur{\'\i}cio~C Magnaguagno, Gal~A Kaminka, and
  Felipe Meneguzzi.
\newblock Towards online goal recognition combining goal mirroring and
  landmarks.
\newblock In {\em AAMAS}, pages 2112--2114, 2018.

\end{thebibliography}

\end{document}